\documentclass{article}
\usepackage{ijcai19}

\usepackage{times}
\usepackage{soul}
\usepackage{url}
\usepackage[hidelinks]{hyperref}
\usepackage[utf8]{inputenc}
\usepackage[small]{caption}
\usepackage{graphicx}
\usepackage{amsmath}
\usepackage{amssymb}
\usepackage{booktabs}
\usepackage{algorithm}
\usepackage{algorithmic}
\usepackage{xcolor}

\title{Explaining Deep Classification of Time-Series Data with Learned Prototypes}

\author{
Alan H. Gee\footnote{Equal Contribution}$^{,1,2}$
\and
Diego Garcia-Olano$^{*,1,2}$\and
Joydeep Ghosh$^1$\And
David Paydarfar$^2$
\affiliations
$^1$Electrical and Computer Engineering, The University of Texas at Austin\\
$^2$Neurology, Dell Medical School, The University of Texas at Austin\\
\emails
\{alangee, diegoolano\}@utexas.edu,
ghosh@ece.utexas.edu,
david.paydarfar@austin.utexas.edu
}

\begin{document}

\maketitle

\begin{abstract}
The emergence of deep learning networks raises a need for explainable AI so that users and domain experts can be confident applying them to high-risk decisions. In this paper, we leverage data from the latent space induced by deep learning models to learn stereotypical representations or ``prototypes" during training to elucidate the algorithmic decision-making process. We study how leveraging prototypes effect classification decisions of two dimensional time-series data in a few different settings: (1) electrocardiogram (ECG) waveforms to detect clinical bradycardia, a slowing of heart rate, in preterm infants, (2) respiration waveforms to detect apnea of prematurity, and (3) audio waveforms to classify spoken digits. We improve upon existing models by optimizing for increased prototype diversity and robustness, visualize how these prototypes in the latent space are used by the model to distinguish classes, and show that prototypes are capable of learning features on two dimensional time-series data to produce explainable insights during classification tasks. We show that the prototypes are capable of learning real-world features - bradycardia in ECG, apnea in respiration, and articulation in speech - as well as features within sub-classes. Our novel work leverages learned prototypical framework on two dimensional time-series data to produce explainable insights during classification tasks.
\end{abstract}

\section{Introduction}
Despite the recent surge of machine learning, adoption of deep learning models in decision critical domains, such as healthcare, has been slow because of limited transparency and explanations in black-box algorithms. This observation points to the critical need for black-box models to offer interpretable, faithful explanations of their decisions so that practitioners in high-risk domains can trust model outputs and leverage their results. One such high-risk domain is treating preterm infants ($\sim$10\% of births worldwide) in the neonatal intensive care unit (NICU).

A common disorder observed in majority of preterm infants is recurrent episodes of apnea (cessation of breathing) and bradycardia (slowing of heart rate). Both of these spontaneous events may cause end organ damage related to hypoxemia (low oxygenation of blood) and ischemia (reduced blood flow) \cite{apnea}. Early detection of apnea and bradycardia can help prevent hypoxic-ischemic injury in tissue with high-metabolic demands \cite{Schmid,Pichler} and prevent the cascade into intermittent hypoxia, which leads to complications of retinopathy, developmental delays, and neuropsychiatric disorders \cite{Williamson,Poets,Fore}. Leveraging explainability in deep neural network classification of these time series can reveal complex morphological and physiological features that clinicians may not readily see. Thus, machine learning algorithms need transparency in their decision-making process to highlight subtle patterns. One such technique in deep explainability is prototypes, a case-based reasoning technique. 

Prototypes are representative examples, learned in-process during model training, that describe influential data regions in latent representations and provide insight into aggregated features across training data that are utilized by the model for classification. In contrast to post-hoc explainability, which trains a secondary model to infer decision reasoning from a primary model by only leveraging inputs and outputs, in-process explainable methods offer faithful explanations of a primary model's decisions \cite{Rudin}. So, users who employ prototypes can confidently gain direct insight into the decisions algorithms are making for classification tasks.  

\begin{figure}[t]
\centering
\includegraphics[width=3.6in]{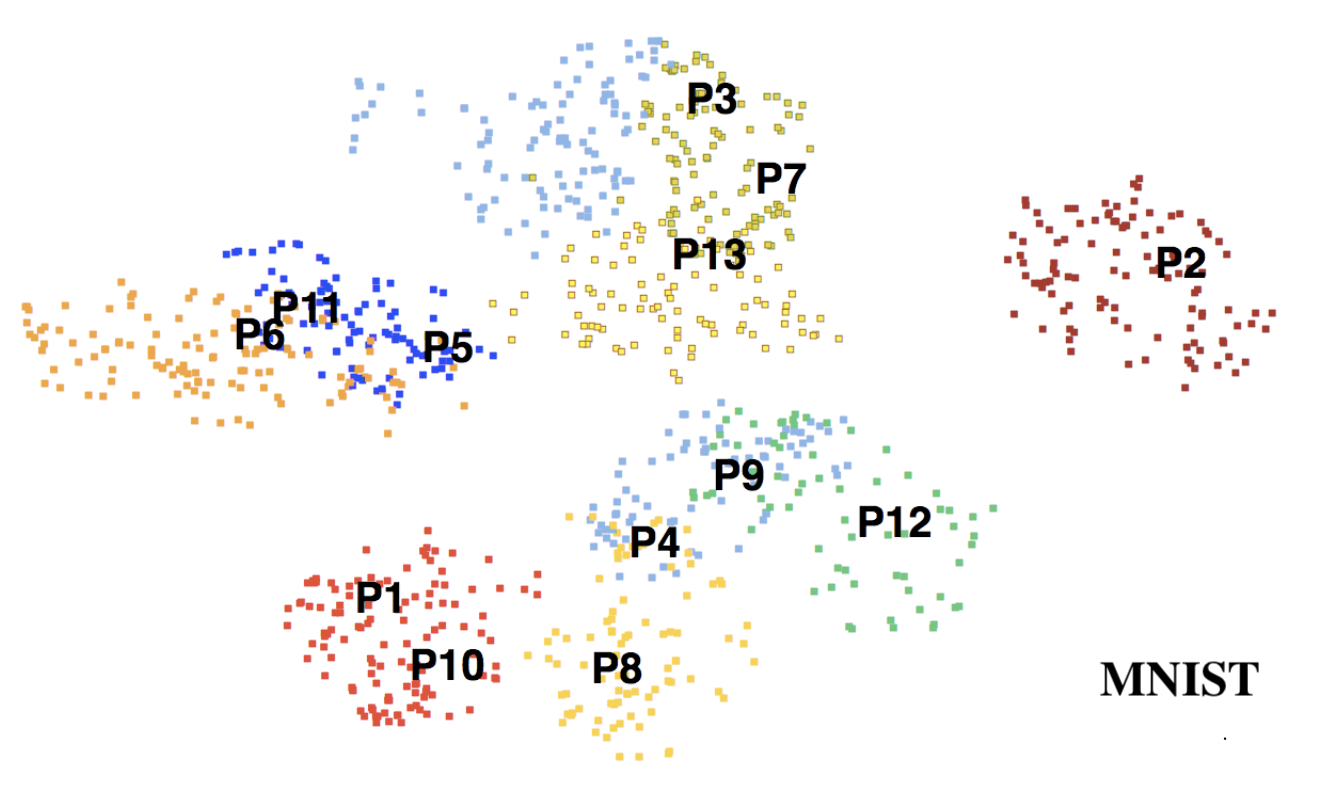}
\caption{Learned prototypes of handwritten digits (MNIST) using the architecture from ~\protect\cite{Li-et-al:proto-2017}. While colors represent the handwritten digits 0-9, the labels represent the learned prototypes. Because the latent representation of MNIST cluster distinctly, the prototypes are diverse. This may not be true when classes overlap} \label{mnist}
\end{figure}

On data with unclear class boundaries, in-process methods can misbehave. For example when the model in ~\cite{Li-et-al:proto-2017} is applied to the MNIST dataset, the prototypes easily separate in the latent space because the latent data representation is separable and well-structured (Fig ~\ref{mnist}). However, when class boundaries and features do not form distinguishable clusters, learned prototypes become archetypes (extreme corner cases) that exist near the convex hull of the data in the latent space (Fig. \ref{fig:regularization}). 
This phenomenon yields prototypes that represent extreme class types (i.e. archetypes) and can underperform on classifying data in overlapping class regions. 

In this work, we provide a deep classification method with explainable insights for health time-series data. We introduce a prototype diversity penalty that explicitly accounts for prototype clustering and encourages the model to learn more diverse prototypes. These diverse prototypes will help focus on areas of the latent space where class separation is most difficult and least defined to improve classification accuracies. We show the utility of this approach on three tasks in two-dimensional time-series classification: (1) bradycardia from ECG; (2) apnea from respiration; and (3) spoken digits from audio waveforms. The two-dimensional representation of time-series provides an interpretable method for domain experts (e.g. clinicians) to understand the evolution of clinically relevant features based on visible phenotypes in time-series data. Our work enables a closed-loop collaboration between  experts and machine learning algorithms to accelerate the efficacy of outcome predictions. The learning algorithms can find nuance features through development of explainable prototypes, and the experts can fine-tune the algorithms by providing feedback through the regularization of the diversity penalty. This is especially important for clinician experts who need explainability in black-box models to understand and diagnose different pathological mechanisms. To the best of our knowledge this is the first application of prototypes and latent space analysis for health time-series data that could help reveal clinically relevant and explainable phenotypes to improve the baseline for standard of care with automatic monitoring and detection.

\subsection{Relevant Work}
Explainable methods \cite{Lime,Caruana,cam} have largely focused on labeled image and tabular data sets where classes are clearly separable and less so on time-series data in general. Recent work has focused on using prototypes to provide in-process explainability of classification models, either by learning meaningful pixels in the entire image ~\cite{Li-et-al:proto-2017} or by applying attention through the use of sub-regions or patches over an image \cite{prototype2}. Class attention maps (CAMs) provide probability maps to highlight areas of images that lead to a certain prediction \cite{cam}, but do not give examples of prototypical examples of the data or explanations of how the training data relates to the end result. We focus on the former work ~\cite{Li-et-al:proto-2017} for example-based explainability where the generation of prototypes are intended to look like global representations of the training data. 

Time-series classification on 1-D data with deep neural networks is a rapidly growing field, with almost 9,000 deep learning models ~\cite{Fawaz,Pons,Faust,Goodfellow}. One such example leverages global average pooling to produce CAMs to provide explainability for a deep CNN to classify atrial fibrillation in ECG data \cite{Goodfellow}. However, the number of available healthcare datasets, specifically ECG waveforms, is limited \cite{Fawaz}. Within this context, time-series classification on ECG waveforms has been done on a small scale, typically with single beat or short-duration (10 s) arrhythmia classification \cite{Faust,Yildirim}.

\section{Methods}
\begin{figure}[t]
  \centering 
  \includegraphics[width=3.45in]{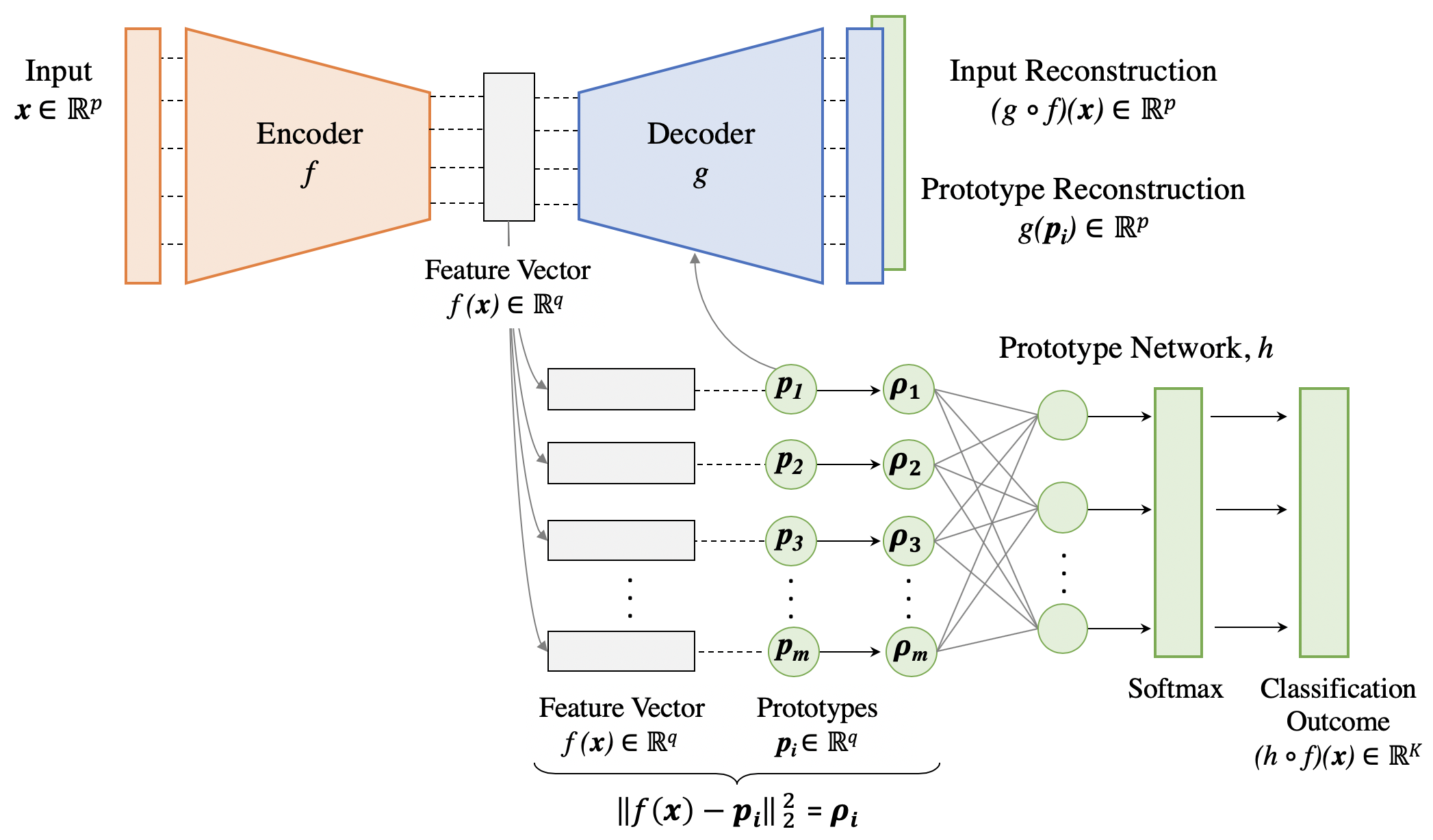}
  \caption{Prototype Architecture from ~\protect\cite{Li-et-al:proto-2017}}
  \label{fig:prototype_arch} 
\end{figure} 

\subsection{Time-Series Explanation via Prototypes}
We adopt the autoencoder-prototype architecture from ~\cite{Li-et-al:proto-2017}. Let $\mathcal{X} = {(x_i, y_i)}_i^n$ be the training set with $x_i \in \mathbb{R}^p$ and class labels $y_i \in \{1,..., K\}$ for each training point $i \in \{1, ..., n\}$. The front-end autoencoder network learns a lower-dimension latent representation of the data with an encoder network, $f: \mathbb{R}^p \rightarrow \mathbb{R}^q$. The latent space is then projected back to the original dimension using a decoder function, $g: \mathbb{R}^q \rightarrow \mathbb{R}^p$. The latent representation, $f(x)$ is also passed to a feed-forward prototype network, $h: \mathbb{R}^q \rightarrow \mathbb{R}^K$, for classification. The prototype network learns $m$ prototype vectors, $p_1, p_2, ..., p_m \in \mathbb{R}^q$ using a four-layer fully-connected network over the latent space that learns a probability distribution over the class labels $y_i$ (Fig \ref{fig:prototype_arch}). The learned prototypes can then be decoded using $g$ and examined to infer what the network has learned.  The choice of $m$ is determined \textit{a priori}, with larger values allowing for higher throughput and model capacity.  

We improve prior work by adding a penalty for learned prototypes in the objective function of the above network to increase prototype diversity and coverage of the data in latent representations. To align with the minimization of the objective function, this new prototype diversity penalty needs to be (1) small when distances between prototypes are far apart, and (2) large when distances between prototypes are close in distance. We can evaluate the feasibility of a set of prototypes by considering the distance of the two closest prototypes across all prototype combinations. So, we consider the average minimum squared $L_2$ distance between any two prototypes, $p_i, p_j$ for our loss function. To achieve the desired property above, we take the inverse of this average distance: 

\begin{equation}
\begin{aligned}
    PDL(p_1, &..., p_m)  =  \\
    & \frac{1}{log \big(\frac{1}{m}\sum_{j=1}^m min_{i > j \in [1,m]} \left\lVert p_i - p_j \right\rVert_2^2 \big) + \epsilon} \label{eq: pd_error}
\end{aligned}
\end{equation}
\newline
The logarithm function tapers large distances so that the penalty does not quickly vanish, and the $\epsilon$ term is for numeric stability. By taking the inverse of the log of the prototype distances, we penalize prototypes that are close in distance while making sure the minimum distance between prototypes does not get too large. This prototype diversity loss (PDL) promotes coverage over the latent space. We update the objective function to:

\begin{equation}
\begin{aligned}
    \mathcal{L}((f, g, h), X) = & \,E( h \circ f, X) + \lambda_R \, R(g \circ f, X) \\ 
    &+ \lambda_1\, R_1 (p_1, ..., p_m, X) \\
    &+ \lambda_2\, R_2(p_1, ..., p_m, X) \\  
    &+ \lambda_{pd} \,PDL(p_1, ..., p_m)  \label{eq:loss}
\end{aligned}
\end{equation}
\newline
where $E$ is the classification (cross entropy) loss, $R$ is the reconstruction loss of the autoencoder (\textit{i.e.} $L_2$ norm), and $R_1$ and $R_2$ are the loss terms that relate the distances of the feature vectors to the prototype vectors in latent space ~\cite{Li-et-al:proto-2017}:

\begin{align}
    R_1(p_1, ..., p_m, X) = & \frac{1}{m} \sum_{j=1}^m min_{i \in [1,n]} \left
    \lVert p_j - f(x_i) \right\rVert_2^2, \\
    R_2(p_1, ..., p_m, X) = & \frac{1}{n} \sum_{i=1}^n min_{j \in [1,m]} \left
    \lVert f(x_i) - p_j \right\rVert_2^2 \label{eq:r_loss}
\end{align}
\newline
The minimization of the $R_1$ loss term promotes each prototype vector to learn one of the encoded training examples, while the minimization of $R_2$ loss promotes encoded training examples to be close to one of the prototypes. This balance gives meaningful pixel-to-pixel representations between the prototypes and training data. 

We train our models with a randomly shuffled batch size of 100 (ECG, Speech) and 125 (Respiration). We parameterize the number of prototypes (see supplement) and the regularization term $\lambda_{pd}$ for the classification tasks while keeping the other hyperparameters as in ~\cite{Li-et-al:proto-2017}. 
\begin{figure}[b!]
  \centering 
  \includegraphics[width=3.38in]{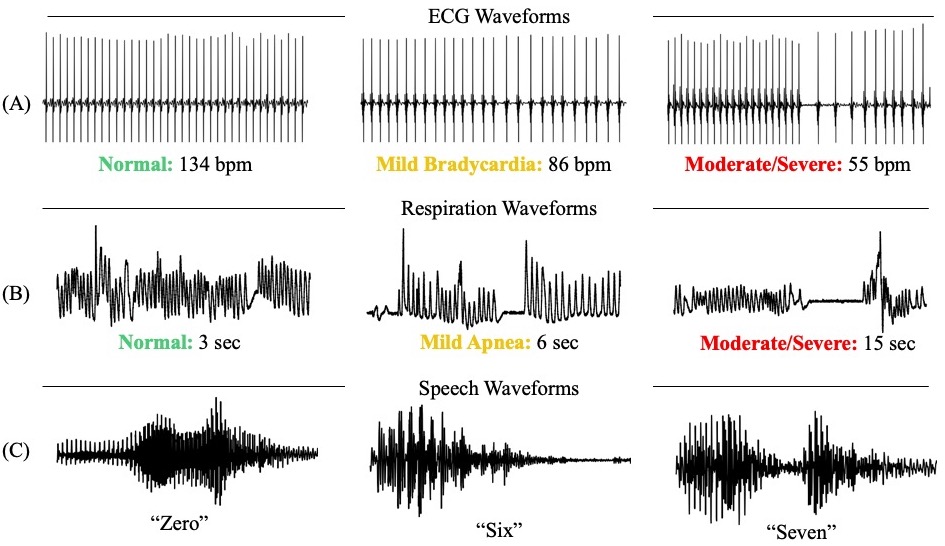} 
  \caption{Examples of waveforms for each task: (A) Electrocardiogram (ECG) waveforms related to bradycardia classification, (B) Respiration waveforms related to apnea classification, and (C) Speech waveforms for a particular a speaker (Jackson). For (A) and (B) we classify the segments based on severity (i.e. time difference between peaks), and for (C) we classify based on digit class.}
  \label{fig:classes} 
\end{figure} 
\subsection{Prototype Diversity Score}
We adopt a version of the group fairness metric presented in \cite{Mehrotra:2018} and refer to it as the prototype diversity score, $\Psi $:
\begin{equation}
    \Psi = \frac{1}{Z} \sum_{i=1}^t \sqrt{ \lvert \phi_i \rvert}
\end{equation}
where $\phi_i, i \in \{1, ..., t\}$ is defined for a specific metric and $Z$ is the normalization constant. For the neighbor diversity metric 
$\Psi_N$, $\phi_i$ is the set of prototypes that have nearest neighbor $i$ and $Z$ is the number of prototypes $m$. For the class diversity metric $
\Psi_C$, $\phi_i$ is the set of prototypes that are from class $i$ and $Z$ is the number of classes $K$. Higher scores will occur when prototypes have more unique elements. Thus, $max(\Psi_D) = 1$.

\subsection{Datasets}
The neonatal intensive care unit (NICU) dataset is composed of two sources: (1) ECG and Respiration waveforms from PhysioNet's PICS database \cite{PIADB,PhysioNet}; and (2) ECG waveforms (500 Hz, Intellivue MP450) collected from a preterm infant over their entire stay ($\sim$10 weeks) at Seton Medical Center Austin. The inclusion of (2) helps supplement the ECG events from (1). The image data used in this study are made publicly available\footnote{https://physionet.org/physiobank/database/picsdb}. 

\begin{figure*}[h]
\centering
\includegraphics[width=6in]{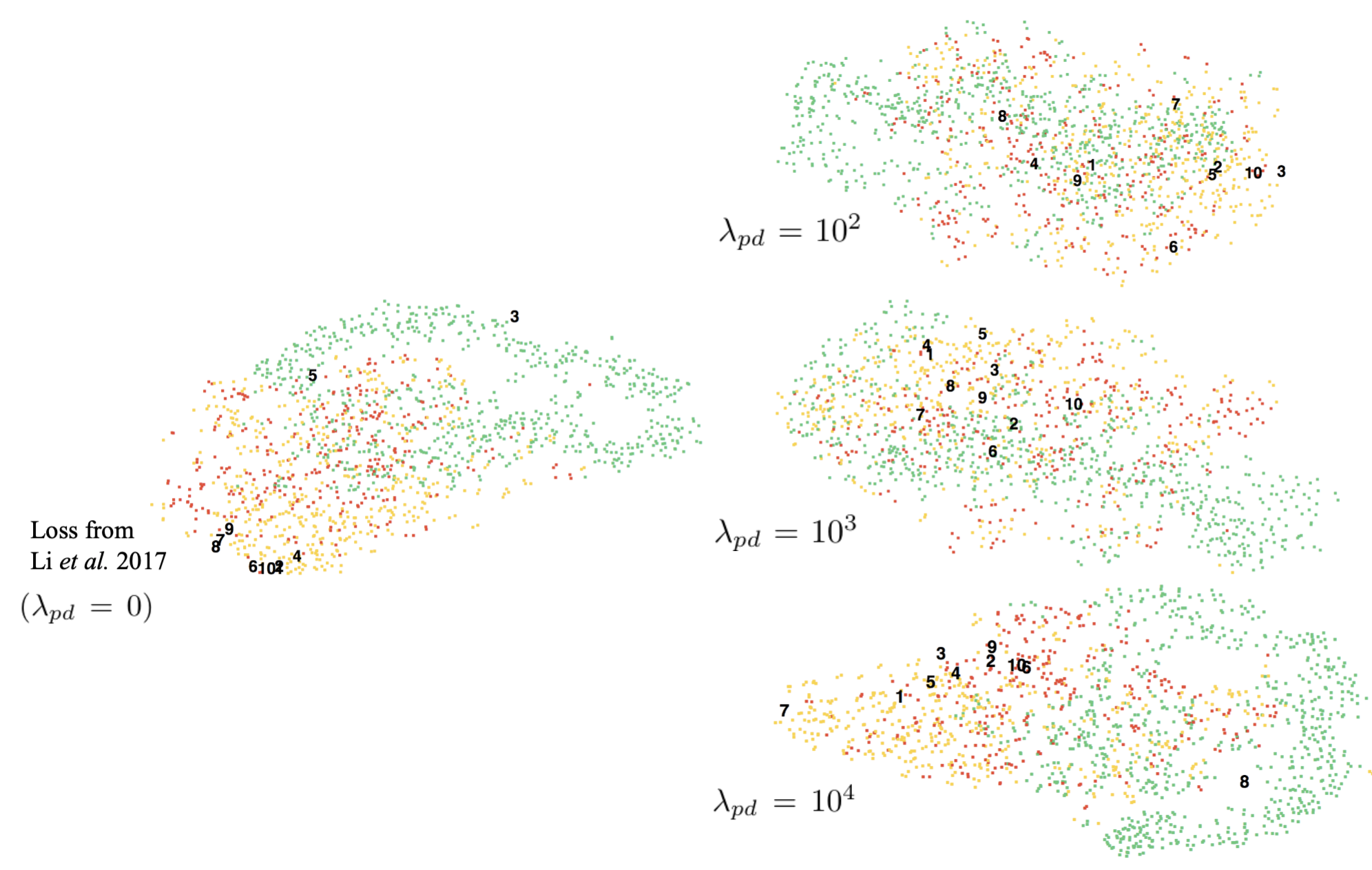}
\caption{Effect of loss regularization on the latent space and spread of prototypes for the NICU classification task using 10 prototypes with $\lambda_{pd} = 0$ (baseline) and $\lambda_{pd} = 10^3$. The second and third dimensions of a t-SNE projection on each space shows prototypes with more coverage and diversity in the latter case.} \label{fig:regularization}
\end{figure*}

The inter-breath intervals (IBIs) from the respiration were extracted using a standard peak finder. The respiration signals were clipped into 60 second segments that were normalized to zero-mean, unit variance. The R-R intervals for the ECG of the NICU dataset were extracted using a Morlet wavelet transformation of the ECG signal. An open-source peak finder was applied to the wavelet scale range (0.01 to .04 scales) related to QRS complex formation in the spectrogram. The ECG waveforms were clipped at 15 seconds with the event in the middle. All ECG segments were band-passed filtered from 3 to 45 GHz, scaled to zero-mean, unit-variance, and scaled to the median QRS complex amplitude. Images were then captured to mimic what a clinician would see upon investigation of an ECG signal. Waveforms with no visibly distinguishable QRS complexes or respiratory peaks were discarded because these waveforms are too obscure for even a clinician expert to evaluate.

Class breakdowns for bradycardia in the ECG signal follow clinical thresholds \cite{Perlman}: $X_{ECG}$ = \{ normal ($>$100 beats per minute (bpm)):  1039, mild (100-80 bpm):  634,  moderate (80-60 bpm): 306,  severe ($<$60 bpm): 132 \}. Moderate and severe events were combined into a single class. The class breakdown for apneas in respiration are: $X_{RESP}$ = \{ normal (1-3 s):  1939, mild (4-6 s):  1921,  moderate/severe ($>$ 6 s): 1487 \}. 

The Free Spoken Digit Dataset \cite{FSDD} consists of 2000 audio clips (8 kHz) of four speakers repeating the digits 0 through 9, 50 times each.  Each segment was normalized to zero-mean, unit-variance and clipped for white space (Fig. \ref{fig:classes}). This data can be thought of as ``spoken MNIST". We perform speaker classification and digit classification within a speaker.

\subsection{Visualization of Latent Space}
 We use PCA to reduce the latent space vectors to a dimension of 500, which retains 98\% of the variability. We then calculate the cosine similarity between these 500 dimensional vectors to produce a similarity matrix and use t-distributed stochastic neighbor embedding (t-SNE) from \cite{tsne} to reduce the 500 x 500 similarity matrix down to three dimensions for visualization purposes. This technique calculates the KL-Divergence between the higher-order dimensional latent space and the lower dimensional space used to represent the former visually. This approach is non-deterministic so the global position in the lower space is uninformative and instead proximity to neighbors is the key insight to gain.  Additionally while the first two dimensions of the projection show the general spread of information, the second and third dimensions maybe useful for visualizing within group information. Thus, we use the second and third dimensions for our visualizations.  

\begin{table*}
\centering
\begin{tabular}{cccc} 
\toprule
\multicolumn{1}{c}{} &     \multicolumn{3}{c}{ECG: Bradycardia}   \\
$\lambda_{pd} $ &  Accu. & $ \Psi_{N}$ &  $ \Psi_{C}$ \\
\midrule
$0$       &  92.1 $\pm$ 0.1\%      &  0.83 $\pm$ 0.04 & 0.78 $\pm$ 0.19  \\
\midrule 
\addlinespace[.2cm]
$500$    &  92.7 $\pm$ 1.0 \%   & 0.86 $\pm$ 0.07 & 0.89 $\pm$ 0.19       \\
$1e3$   & 92.4 $\pm$ 1.3\%   & 0.87 $\pm$ 0.11 & 0.89 $\pm$ 0.19     \\
$2e3$   & \bf{93.1 $\pm$ 0.4\%} & \bf{0.90 $\pm$ 0.04} & \bf{1.00 $\pm$ 0.00}   \\
\bottomrule
\end{tabular} \qquad
\begin{tabular}{cccc}  
\toprule
 \multicolumn{1}{c}{} &    \multicolumn{3}{c}{Respiration: Apnea}   \\
$\lambda_{pd} $ & Acc. & $\Psi_{N}$ &  $ \Psi_{C}$  \\
\midrule
 $0$ & 81.4 $\pm$ 3.6\%   &  0.96 $\pm$ 0.07 & \bf{1.00 $\pm$ 0.00}   \\
  \midrule 
\addlinespace[.2cm]
$500$ &  \bf{82.3 $\pm$ 3.8\%}   & 0.94 $\pm$ 0.09 & \bf{1.00 $\pm$ 0.00}       \\
$1e3$ & 77.1 $\pm$ 0.6 \%& \bf{ 1.00 $\pm$ 0.00}  &  \bf{1.00 $\pm$ 0.00}       \\
$2e3$ &  80.2 $\pm$ 2.5 \%   & 0.97 $\pm$ 0.04 & 0.84 $\pm$ 0.23      \\

\bottomrule
\end{tabular}
  \caption{Diversity score for neighbors $\Psi_N$ and class $\Psi_C$. We report $\Psi$'s related to the epoch with the highest test accuracy. Our model, $\lambda_{pd} > 0$, returns better accuracies and diversity scores (bold) than the baseline model, which is row $\lambda_{pd} = 0$, across ECG and Respiration waveforms. (Model details: \textit{3-class, 10-prototypes, learning rate = 0.002}).}\label{tab1}
\end{table*}

\section{Results}
\subsection{Classification of ECG with 2-D Prototypes}
We test our prototype implementation on ECG waveforms related to bradycardia using the NICU data for a 3-class classification task using 10 prototypes.  We treat the input waveforms as 2-D images and use a four-layer autoencoder to learn complex representations over the data.  

We observe more diverse prototypes and comparable or better test accuracy with our model 93.1$\pm$0.4\% compared with 92.1$\pm$0.1\% from the baseline model in ~\cite{Li-et-al:proto-2017} (Table \ref{tab1}). Both models perform well on the classification of the normal class, as expected since normal waveforms have near-constant phase. Both models additionally have difficulty separating between the mild and moderate/severe classes, often confusing the classification between these two (see supplement). This behavior is expected since data near these two class boundaries are difficult to discern, even for domain experts, due to events existing in both classes with possible subtle time differences in cardiac firing. Our model also improves prototype diversity (Table \ref{tab1}) over the baseline model. This result suggests that the prototype diversity loss encourages exploration, through learning diverse prototypes, within the data represented in the latent space. As a result, our model finds more helpful features and prototypes and thus, improves classification results.

\begin{figure}[h]
\centering
\includegraphics[width=3.47in]{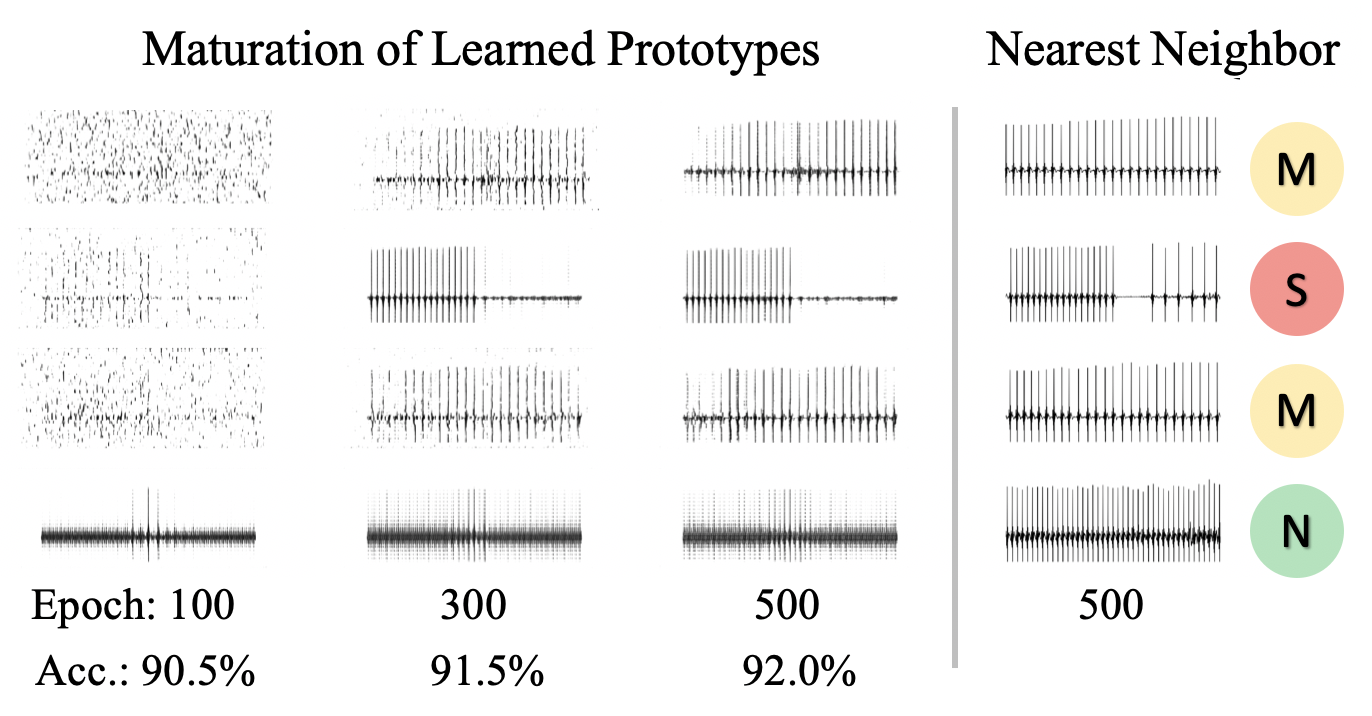}
\caption{Prototype evolution with in-process explainability over training time. High level features are easily learned in early epochs of training, while more complex features are developed over time. The final nearest neighbors are depicted on the right. The prototypes correspond to a subset of the $\lambda_{pd} = 10^3$ latent space cloud in Figure \ref{fig:regularization}. \textit{Model details: 3-class, 10-prototypes}.} \label{evolution}
\end{figure}

Because prototypes are generated during training, we infer features that the algorithm utilized to classify waveforms at different points during training (Fig \ref{evolution}). For example, by epoch 100, we see that some of the prototypes exhibit global morphological features of the normal waveform class after random initialization at epoch 0. As training progresses, we observe other complex phenotypes emerging: one prototype learns that large gaps in cardiac firings are important for identifying severe cases and another prototype learns the consistent pattern of spikes are important for mild cases. Since the mild class shares mixed features of both normal and positive events, it is not surprising that more prototypes are needed in this class to learn subtleties of the class features (see supplement). Thus, prototypes highlight waveform structures that the algorithm deemed as important when trying to learn the classification of bradycardia. This finding aligns with the idea of clinicians using visible features present in a bradycardia (i.e. the increasing distance between QRS complexes) to decide whether or not a bradycardia exists in an image.

We compare the latent space of ~\cite{Li-et-al:proto-2017} to the latent space of our model with prototype diversity loss via t-SNE projections, where proximity in 2-D space suggests that points are ``close" in distance in the original latent space. We represent the learned prototypes by mapping each prototype to its nearest neighbor (Fig \ref{fig:regularization}). We find that by increasing our loss term, $PDL$, our model increases the local coverage of the prototypes compared with the baseline model (i.e. $\lambda_{pd} = 0$). However, if we regularize our loss term too much (i.e. $\lambda_{pd} > 10^4$), we begin to introduce clustering of prototypes and diversity suffers.  Thus with the additional prototype distance penalty, we achieve higher diversity scores and classification accuracies for various hyperparameters (Fig \ref{fig:accuracy}). 

 \begin{figure*}[h!]
\centering
\includegraphics[width=4.75in]{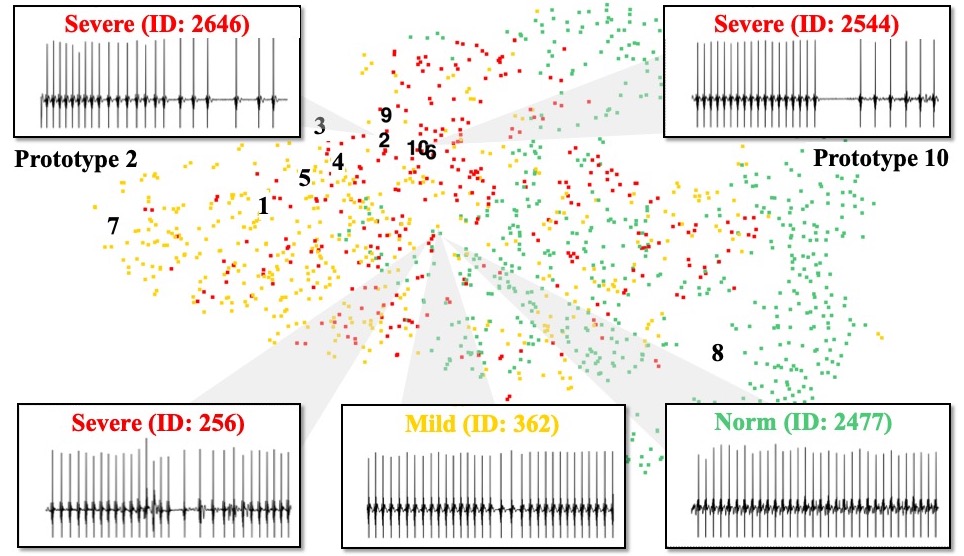}
\caption{Learned prototypes showcase the diversity of features that are important for understanding ECG morphology while classifying bradycardia events. \textit{( 10-prototypes, $\lambda_{pd} = 10^4$} ).} \label{fig:morphology}
\end{figure*}

\subsection{Case Study with Prototypes: Exploring ECG Morphology and Bradycardia Classification.} 
We observe that ECG events in a local neighborhood share similar QRS complex morphology, despite having different class labels and cardiac firing periods (Fig. \ref{fig:morphology}, bottom). Even though we did not impose a class constraint, we observe that the algorithm found two separate features within the moderate/severe class that were important in the classification task (i.e. prototypes 2 and 10 shown at the top of the (Fig \ref{fig:morphology}). These two prototypes explore two different cardiac timings as prototype 2 exhibits a progressive delay in cardiac firing, while prototype 10 exhibits a large spontaneous delay. The incorporation of the prototype diversity loss encouraged this exploration of the latent space. These results suggest that there are physiologic dependencies (i.e. clustering based on cardiac morphology and function) that can be learned using our model to investigate physiological phenomena, and possibly applied to other clinical areas, like cardiac ischemia or apnea of prematurity in respiration - both exhibit visible, abnormal waveform behavior. This work provides a visualization tool for clinician experts to evaluate different morphological of physiological time-series data\footnote{https://github.com/alangee/ijcai19-ts-prototypes}.

\begin{figure}[h]
\centering
\includegraphics[width=3.36in]{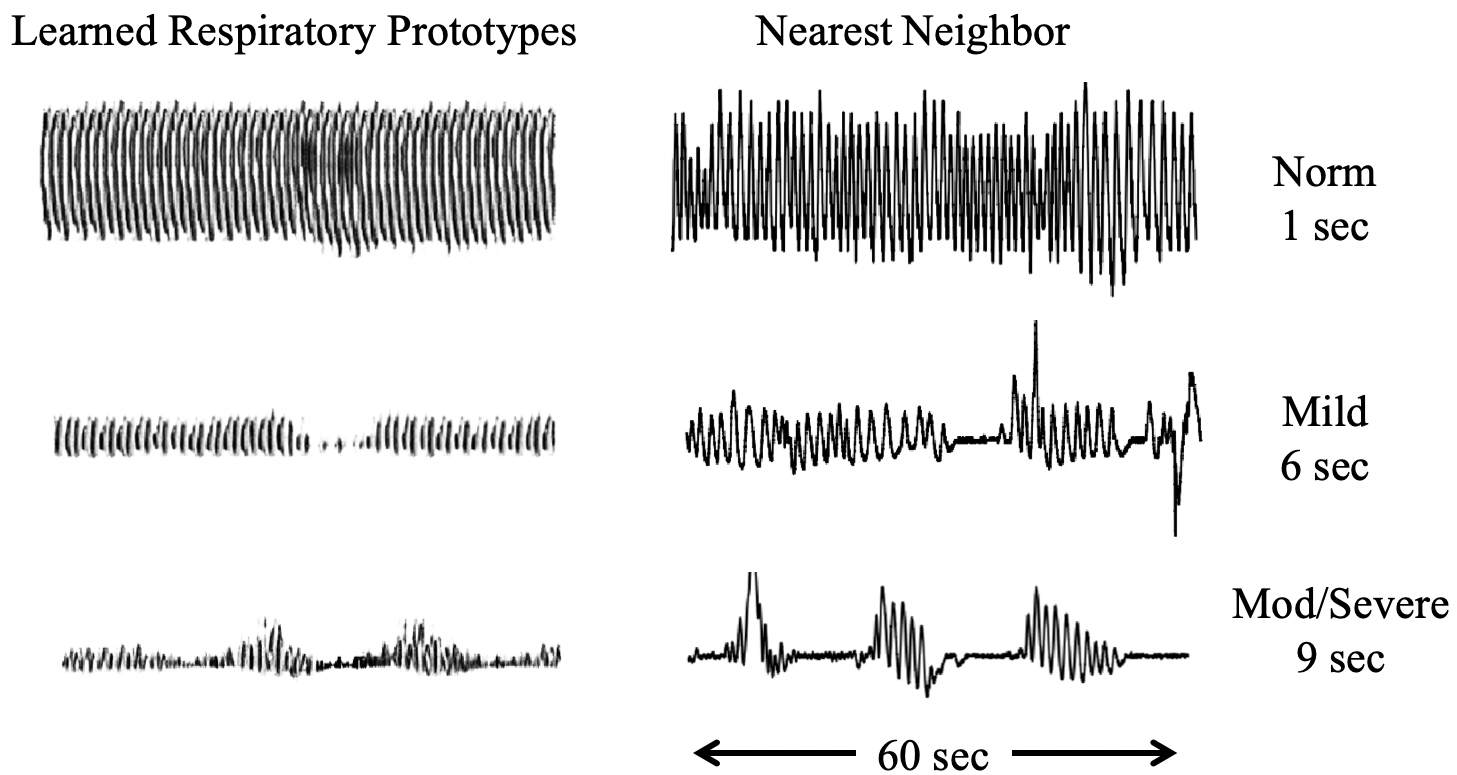}
\caption{Learned prototypes showcase the diversity of features across classes that are important for understanding respiration morphology while classifying apnea events. For this classification task, we observe a variety of prototypes (at epoch 500) that learn various cases with cessation of breathing (6 and 9 second gaps) and the global features within the segment that are important for the model's classification. \textit{(8-prototypes, $\lambda_{pd} = 500$}).} 
\label{fig:respiratory_protos}
\end{figure}

\begin{figure}[h]
\centering
\includegraphics[width=3.25in]{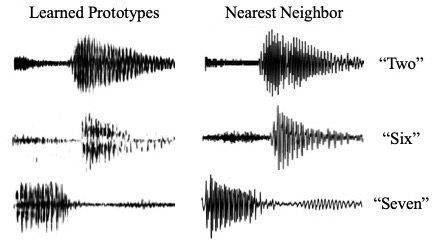}
\caption{Learned prototypes from audio waveforms of spoken digits by Nicolas from the FSDD \textit{ ($\lambda_{pd} = 500$)}.} \label{fig:mnist}
\end{figure}

\subsection{Classification of Apnea in Respiration}
Apnea of prematurity is common among preterm infants, and is visually apparent as a pause of inhalation and exhalation (i.e. absence of sinusoidal behavior) in the respiratory signal. We next test our prototype implementation on respiration waveforms that are related to apnea in a 3-class classification task. We treat the input waveforms as 2-D images again, since clinicians evaluate apneas through visual inspection of the respiration signal.

We observe more diverse prototypes and comparable or better test accuracy with our model 82.3$\pm$3.8\% compared with 81.4$\pm$3.6\% from the baseline model, and with overall unique nearest neighbors ($\Psi_N$ = 1) and class diversity ($\Psi_C$ = 1) (Table \ref{tab1}). Both models have difficulty separating between the event classes because data near these two class boundaries are difficult to visually discern (i.e. 6 second gap versus 7 second gap) and have common behavior with regular respiratory function that is found in the normal class. We find that the addition of a prototype diversity loss maintains or improves performance and yields more diverse prototypes (Table \ref{tab1}). 

We also note that the algorithm is able to discern physiological examples and generate learned prototypes that distinctly relate to physiological behavior. For example, in Fig. \ref{fig:respiratory_protos}, we see that algorithm finds segments that are related to periodic breathing of 9 second duration (moderate/severe). These segments are physiologically different from normal apneas of 6 seconds (mild), and clearly different from normal breathing with periodicity of 1 second (Fig \ref{fig:respiratory_protos}). In the set of eight learned prototypes, the algorithm finds three different classes easily, each with different respiratory phenomena, that are critical in the classifying various types of apneas. 

\subsection{Spoken Digits Classification and Analysis}
\begin{figure}
\centering
\includegraphics[width=3.35in]{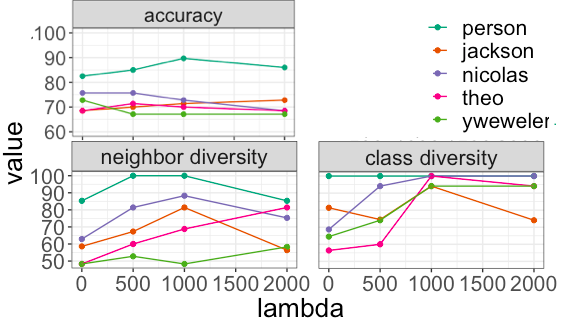}
\caption{Accuracy and diversity metrics for the spoken digits experiments using the FSDD. We divide this dataset into two tasks: (1) classifying the person speaking and (2) classifying the digit spoken within each person. } \label{fig:accuracy}
\end{figure}

Speech abnormalities can be suggestive of underlying pathological dysfunction, and common features that clinicians visibly discern in waveforms to assess speech include cadence, prosody, and syllable articulation. To aid in speech feature detection, we assess our model on high-frequency audio waveforms of spoken digits (FSDD) from medically-normal individuals. These digits are treated as 2-D images for 4 class speaker and 10 digit classification tasks with 4 and 10 prototypes, respectively. The waveform envelope and syllables of these spoken digits are discernible to the eye (see \textit{``six"} and \textit{``se-ven"} in Fig 2) and, as such, make good candidates for our image-based explainability model. We demonstrate some of the learned prototypes in Fig. \ref{fig:mnist}, which show representations the model finds useful in classifying digits for a given speaker. Experiments show that by varying regularization of the prototype diversity penalty, we observe slightly better or similar accuracies when compared to the baseline model (Fig. \ref{fig:accuracy}). With a fine-tuned $\lambda_{pd}$ we can increase diversity of the prototypes and correspondingly see improved accuracy and data coverage (see supplement). For example, $\lambda_{pd} = 500$ gives a higher diversity score across all tasks, indicating prototypes with more unique nearest neighbors as compared with the baseline model (Fig \ref{fig:accuracy}). 

Experiments show that increasing the depth of the network and fine-tuning the learning rate lead to both increased accuracy and diversity over all tasks. Similarly, recent data augmentation techniques in medical \cite{bahadori2019temporalclustering} and speech recognition \cite{park2019specaugment} domains could help further improve performance.  The purpose of this work, however, is not to obtain the best performance on these tasks, but rather to show the utility of learned prototypes as faithful explanations of decisions made by a model. 

\section{Discussion} 
We presented a new autoencoder-prototype model that promotes diversity in learned prototypes by penalizing prototypes that are too close in squared $L_2$ distance in the latent space. The new term, $\lambda_{pd} \, PDL(p_1, ..., p_m)$, in the loss function (Eq. \ref{eq:loss}) promotes prototype diversity while improving classification accuracy and prototype coverage of data represented in the latent space.  These prototypes help explain which global features and representative segments in the training data are most useful for deep time-series classification. This in-process generation of prototypes offers explainable insights into deep classifiers.

Our model and results provide an important significance that previous works lack. Depending on the clinical context of the case, experts may want to either trivialize big differences in the time series features, or conversely accentuate nuanced differences in learned prototypes as clinically important signs of impending adverse outcomes. Therefore, our implementation offers a collaborative method for clinician experts to use their insight interactively with machine learning algorithms: increasing $\lambda_{pd}$ promotes large observable differences in the prototypes, while decreasing $\lambda_{pd}$ promotes diverse features and prototypes. In turn, our model enables a closed-loop feedback framework to accelerate phenotype discovery to lead clinicians to better-informed decision.

We evaluate the performance of our model on increasingly difficult physiological datasets to demonstrate the effect of $\lambda_{pd}$. The ECG signal is more robust against movement artifact and produces a cleaner signal for the 2-D visualization task, whereas the respiration signal, which is the resultant voltage change across diaphragm movement, is highly susceptible to signal artifact. Additionally, speech waveforms are compressed, high-frequency waveforms (kHz) which make it difficult to visibly extract high-resolution features. We find that our model allocates more prototypes to learn the intricacies of the more indistinguishable classes (i.e. mild and moderate/severe) that are hard for a human to discern, especially the mild cases because this class is a mixture and intermediary of the two extreme classes. 

We observe, however, that the high number of loss terms creates a trade-off between prototype interpretability and model accuracy. For example, we observe that for a small number of prototypes, we achieve near-perfect prototype reconstruction but at the cost of classification accuracy. When the number of prototypes was large, we achieve higher accuracy but received noisy prototypes. In future implementations, we can replace the front-end autoencoder with a model that operates well on 1-D time series, like an recurrent neural network, to balance accuracy and prototype interpretability.

There has also been work on computing prototypical patches over 2-D images to generate explainable sub-features \cite{prototype2}. Extending the idea of patches to 1-D time-series signals would allow for parsing the signal for sub-frequencies and features that could better explain how events are triggered. Nonetheless, the work presented in this paper provides a more robust prototype model to help explain algorithmic behavior and decision-making in deep time-series classification tasks with promising results in clinically relevant datasets.

\section*{Acknowledgement}
The authors would like to thank Sinead Williamson and the three reviewers for providing helpful feedback and critical reviews of our work.

\bibliographystyle{named}
\bibliography{proto}

\begin{thebibliography}{}

\bibitem[\protect\citeauthoryear{Bahadori and
  Lipton}{2019}]{bahadori2019temporalclustering}
Mohammad~Taha Bahadori and Zachary~Chase Lipton.
\newblock Temporal-clustering invariance in irregular healthcare time series.
\newblock {\em arXiv preprint arXiv:1904.12206}, 2019.

\bibitem[\protect\citeauthoryear{Caruana \bgroup \em et al.\egroup
  }{2015}]{Caruana}
Rich Caruana, Yin Lou, Johannes Gehrke, Paul Koch, Marc Sturm, and Noemie
  Elhadad.
\newblock Intelligible models for healthcare: Predicting pneumonia risk and
  hospital 30-day readmission.
\newblock In {\em Proceedings of the 21th ACM SIGKDD International Conference
  on Knowledge Discovery and Data Mining}, KDD '15, pages 1721--1730, New York,
  NY, USA, 2015. ACM.

\bibitem[\protect\citeauthoryear{Chen \bgroup \em et al.\egroup
  }{2018}]{prototype2}
Chaofan Chen, Oscar Li, Alina Barnett, Jonathan Su, and Cynthia Rudin.
\newblock This looks like that: deep learning for interpretable image
  recognition.
\newblock {\em CoRR}, abs/1806.10574, 2018.

\bibitem[\protect\citeauthoryear{Di~Fiore \bgroup \em et al.\egroup
  }{2015}]{Fore}
J.M. Di~Fiore, E~Gauda, R.J. Martin, and P~MacFarlane.
\newblock Cardiorespiratory events in preterm infants: interventions and
  consequences.
\newblock {\em Journal Of Perinatology}, 36(251), 2015.

\bibitem[\protect\citeauthoryear{Faust \bgroup \em et al.\egroup
  }{2018}]{Faust}
Oliver Faust, Yuki Hagiwara, Tan~Jen Hong, Oh~Shu Lih, and U~Rajendra Acharya.
\newblock Deep learning for healthcare applications based on physiological
  signals: A review.
\newblock {\em Computer Methods and Programs in Biomedicine}, 161:1 -- 13,
  2018.

\bibitem[\protect\citeauthoryear{Fawaz \bgroup \em et al.\egroup
  }{2018}]{Fawaz}
Hassan~Ismail Fawaz, Germain Forestier, Jonathan Weber, Lhassane Idoumghar, and
  Pierre{-}Alain Muller.
\newblock Deep learning for time series classification: a review.
\newblock {\em CoRR}, abs/1809.04356, 2018.

\bibitem[\protect\citeauthoryear{Gee \bgroup \em et al.\egroup }{2017}]{PIADB}
A.~H. Gee, R.~Barbieri, D.~Paydarfar, and P.~Indic.
\newblock Predicting bradycardia in preterm infants using point process
  analysis of heart rate.
\newblock {\em IEEE Transactions on Biomedical Engineering}, 64(9):2300--2308,
  2017.

\bibitem[\protect\citeauthoryear{Goldberger \bgroup \em et al.\egroup
  }{2000}]{PhysioNet}
Ary~L. Goldberger, Luis A.~N. Amaral, Leon Glass, Jeffrey~M. Hausdorff,
  Plamen~Ch. Ivanov, Roger~G. Mark, Joseph~E. Mietus, George~B. Moody,
  Chung-Kang Peng, and H.~Eugene Stanley.
\newblock {PhysioBank}, {PhysioToolkit}, and {PhysioNet}: Components of a new
  research resource for complex physiologic signals.
\newblock {\em Circulation}, 101(23):e215--e220, June 2000.

\bibitem[\protect\citeauthoryear{Goodfellow \bgroup \em et al.\egroup
  }{2018}]{Goodfellow}
Sebastian Goodfellow, Andrew Goodwin, Danny Eytan, Robert Greer, Mjaye Mazwi,
  and Peter Laussen.
\newblock Towards understanding ecg rhythm classification using convolutional
  neural networks and attention mappings.
\newblock In {\em Proceedings of Machine Learning for Healthcare}, MLHC '18,
  pages 2243--2251, 08 2018.

\bibitem[\protect\citeauthoryear{Jackson \bgroup \em et al.\egroup
  }{2018}]{FSDD}
Zohar Jackson, C{\'e}sar Souza, Yuxin Flaks, Jason;~Pan, Hereman Nicolas, and
  Adhish Thite.
\newblock Free spoken digit dataset (fsdd).
\newblock 2018.

\bibitem[\protect\citeauthoryear{Li \bgroup \em et al.\egroup
  }{2017}]{Li-et-al:proto-2017}
Oscar Li, Hao Liu, Chaofan Chen, and Cynthia Rudin.
\newblock Deep learning for case-based reasoning through prototypes: A neural
  network that explains its predictions.
\newblock {\em CoRR}, abs/1710.04806, 2017.

\bibitem[\protect\citeauthoryear{Martin and Wilson}{2012}]{apnea}
Richard~J. Martin and Christopher~G. Wilson.
\newblock Apnea of prematurity.
\newblock pages 2923--2931, 2012.

\bibitem[\protect\citeauthoryear{Mehrotra \bgroup \em et al.\egroup
  }{2018}]{Mehrotra:2018}
Rishabh Mehrotra, James McInerney, Hugues Bouchard, Mounia Lalmas, and Fernando
  Diaz.
\newblock Towards a fair marketplace: Counterfactual evaluation of the
  trade-off between relevance, fairness \& satisfaction in recommendation
  systems.
\newblock In {\em Proceedings of the 27th ACM International Conference on
  Information and Knowledge Management}, pages 2243--2251, 2018.

\bibitem[\protect\citeauthoryear{Park \bgroup \em et al.\egroup
  }{2019}]{park2019specaugment}
Daniel~S Park, William Chan, Yu~Zhang, Chung-Cheng Chiu, Barret Zoph, Ekin~D
  Cubuk, and Quoc~V Le.
\newblock Specaugment: A simple data augmentation method for automatic speech
  recognition.
\newblock {\em arXiv preprint arXiv:1904.08779}, 2019.

\bibitem[\protect\citeauthoryear{Perlman and Volpe}{1985}]{Perlman}
Jeffrey~M. Perlman and Joseph~J. Volpe.
\newblock Episodes of apnea and bradycardia in the preterm newborn: Impact on
  cerebral circulation.
\newblock {\em Pediatrics}, 76(3):333--338, 1985.

\bibitem[\protect\citeauthoryear{Pichler \bgroup \em et al.\egroup
  }{2003}]{Pichler}
G.~Pichler, B.~Urlesberger, and W.~Muller.
\newblock Impact of bradycardia on cerebral oxygenation and cerebral blood
  volume using apnoea in preterm infants.
\newblock {\em Physio. Measurement}, 24(3):671--680, 2003.

\bibitem[\protect\citeauthoryear{Poets \bgroup \em et al.\egroup
  }{2015}]{Poets}
Christian~F. Poets, Robin~S. Roberts, Barbara Schmidt, Robin~K. Whyte,
  Elizabeth~V. Asztalos, David Bader, Aida Bairam, Diane Moddemann, Abraham
  Peliowski, Yacov Rabi, Alfonso Solimano, and Harvey Nelson.
\newblock Association between intermittent hypoxemia or bradycardia and late
  death or disability in extremely preterm infants.
\newblock {\em JAMA}, 314(6):595--603, 08 2015.

\bibitem[\protect\citeauthoryear{Pons \bgroup \em et al.\egroup }{2017}]{Pons}
Jordi Pons, Oriol Nieto, Matthew Prockup, Erik~M. Schmidt, Andreas~F. Ehmann,
  and Xavier Serra.
\newblock End-to-end learning for music audio tagging at scale.
\newblock {\em CoRR}, abs/1711.02520, 2017.

\bibitem[\protect\citeauthoryear{Ribeiro \bgroup \em et al.\egroup
  }{2016}]{Lime}
Marco~T{\'{u}}lio Ribeiro, Sameer Singh, and Carlos Guestrin.
\newblock "why should {I} trust you?": Explaining the predictions of any
  classifier.
\newblock {\em CoRR}, abs/1602.04938, 2016.

\bibitem[\protect\citeauthoryear{Rudin}{2018}]{Rudin}
Cynthia Rudin.
\newblock Please stop explaining black box models for high stakes decisions.
\newblock {\em CoRR}, abs/1811.10154, 11 2018.

\bibitem[\protect\citeauthoryear{Schmid \bgroup \em et al.\egroup
  }{2015}]{Schmid}
M.B. Schmid, R.J. Hopfner, S.~Lenhof, H.D. Hummler, and H.~Fuchs.
\newblock Cerebral oxygenation during intermittent hypoxemia and bradycardia in
  preterm infants.
\newblock {\em Neonatology}, 107:137--146, 2015.

\bibitem[\protect\citeauthoryear{Van~der Maaten and Hinton}{2008}]{tsne}
L.~Van~der Maaten and G.~Hinton.
\newblock Visualizing data using t-sne.
\newblock {\em Journal of Machine Learning Research}, 9:2579--2605, 2008.

\bibitem[\protect\citeauthoryear{Williamson \bgroup \em et al.\egroup
  }{2013}]{Williamson}
James~R. Williamson, Daniel~W. Bliss, and David Paydarfar.
\newblock Forecasting respiratory collapse: Theory and practice for averting
  life-threatening infant apneas.
\newblock {\em Respiratory Physiology \& Neurobiology}, 189(2):223 -- 231,
  2013.

\bibitem[\protect\citeauthoryear{Yildirim \bgroup \em et al.\egroup
  }{2018}]{Yildirim}
Ozal Yildirim, Pawel Plawiak, Ru-San Tan, and U.~Rajendra Acharya.
\newblock Arrhythmia detection using deep convolutional neural network with
  long duration ecg signals.
\newblock {\em Computers in Biology and Medicine}, 102:411 -- 420, 2018.

\bibitem[\protect\citeauthoryear{Zhou \bgroup \em et al.\egroup }{2015}]{cam}
Bolei Zhou, Aditya Khosla, {\`{A}}gata Lapedriza, Aude Oliva, and Antonio
  Torralba.
\newblock Learning deep features for discriminative localization.
\newblock {\em CoRR}, abs/1512.04150, 2015.

\end{thebibliography}

\end{document}